\documentclass[10pt,letterpaper]{article}

\usepackage{amssymb, amsbsy}
\usepackage{cogsci}
\usepackage{graphicx}
\usepackage{hhline}
\usepackage{makecell}
\usepackage{multirow}
\usepackage{pifont}
\usepackage{threeparttable}

\cogscifinalcopy 

\usepackage[
  style=apa,
  natbib=true,
  annotation=false,
]{biblatex}
\usepackage{float} 

\usepackage{marvosym} 

\title{Knowledge-informed Bidding with Dual-process Control for Online Advertising}

\author{
 {\large\bfseries Huixiang Luo$^1$, Longyu Gao$^1$, Yaqi Liu$^2$, Qianqian Chen$^1$, Pingchun Huang$^2$, Tianning Li$^1$\textsuperscript{\Letter}} \\
{\normalsize\normalfont
$^1$ Alibaba Health, Hangzhou, China\\
$^2$ Alibaba Health, Beijing, China
}\\
\texttt{\{luohuixiang.lhx, longyu.gly, fangqi.cqq, tianning.litn\}@alibaba-inc.com}\\
\texttt{\{lyq418445, pingchun.hpch\}@alibaba-inc.com}
}

\begin{document}

\maketitle

\begin{abstract}
Bid optimization in online advertising relies on black-box machine learning models that learn bidding decisions from historical data.
However, these approaches sometimes fail to replicate human experts’ adaptive, experience-driven, and globally coherent decisions.
Specifically, they generalize poorly in data-sparse cases because of missing structured knowledge, make short-sighted decisions that ignore long-term interdependencies, and struggle to adapt in out-of-distribution scenarios where human experts succeed.
To address this, we propose KBD (Knowledge-informed Bidding with Dual-process control), a novel method for bid optimization. 
KBD embeds human expertise as inductive biases through the Informed Machine Learning paradigm, uses Decision Transformer (DT) to globally optimize multi-step bidding sequences, and implements dual-process control by combining a fast rule-based PID (System 1) with DT (System 2).
Extensive experiments highlight KBD’s advantage over existing methods and underscore the benefit of grounding bid optimization in human expertise and dual-process control.

\textbf{Keywords:}
Informed Machine Learning;
Dual-process Theory;
Decision Transformer;
Auto-bidding;
Online Advertising
\end{abstract}

\section{Introduction}

Auto-bidding now dominates digital impression purchasing, projected to exceed \textdollar 200 billion dollar in ad spend in 2026 and to comprise over 92\% of online ad transactions \citep{emarketer2025programmaticH1}.
In ad platforms, advertisers choose appropriate auto-bidding strategies and specify high-level performance goals (e.g., target Return-on-Investment (tROI), target Cost-Per-Action (tCPA), and target Cost-Per-Click (tCPC)) for ad campaigns, while the platform handles impression procurement by bidding automatically for each ad impression under advertisers' guidance. 
Since performance goals are advertisers’ main control lever and directly affect revenues (e.g., Gross Merchandise Volume, GMV), setting goal decisions effectively is crucial for achieving procurement objectives.
Prior works \citep{castiglioni2025safe, liang2023online, luo2024puros} seek to replace human experts with black-box Machine Learning (ML) models that learn bid optimization policies from historical data, enabling scalable bid optimization system. 
However, such models often underperform experts in data-scarce settings, optimize myopically at a single decision step while ignoring inter-temporal interactions, and generalize poorly to abrupt distribution shifts (e.g., sales promotions and new-product launches).

To address these limitations, we propose KBD (Knowledge-informed Bidding with Dual-process Control), a two-stage bid optimization method. 
At the macro (daily) stage, we adopt Informed Machine Learning (IML) paradigm \citep{von2021informed} to embed human expertise into model across hypothesis, algorithm and data levels. 
The hypothesis level design is central to macro stage bid optimization. 
We construct a price-volume model with hybrid cognitive architectures \citep{mumuni2024improving} to produce base bid values that are robust to data sparsity.
At the micro (hourly) stage, we introduce Decision Transformer (DT) \citep{chen2021decision} to optimize multi-step future rewards, mitigating the short-sightedness of step-wise optimization.
To handle abrupt distribution shifts, we draw on dual-process control \citep{moskovitz2022unified}: the rule-based PID controller \citep{bennett1993development} (System 1) guides the complicated DT model training (System 2), and their outputs are combined to improve robustness in extreme scenarios.

In summary, our contributions are threefold: 

\noindent(1) We propose KBD, a two stage bid optimization method that couples expert-driven daily calibration with sequential hourly control to optimize long-term bidding rewards.

\noindent(2) We improve model performance and robustness with a dual-process controller, where PID (System 1) guides and is fused with DT (System 2) to handle data distribution shifts.

\noindent(3) Experiments on two datasets show the effectiveness and broad applicability of KBD across diverse environments.

\section{Related Works}

\subsection{Auto-bidding Strategies in Ad Platforms}

Ad platforms offer various auto-bidding strategies to meet advertisers’  needs for ad procurement. 
These strategies eliminate manual per-impression bidding, substantially reducing the complexity of ad procurement and enabling advertisers to get started with minimal effort.
However, all these black-box strategies also aim to maximize platforms' advertising revenue, so low-value impressions are always mixed with high-value impressions and sold together to advertisers.
As a result, advertisers often adapt to or strategically circumvent these platform-controlled strategies.
A common approach is to calibrate performance goals to optimize bids, thereby reducing low-value impression purchases and improving ad performance.
Mainstream auto-bidding strategies fall into three groups: (1) rule-based controllers, e.g., RTBControl \citep{weinan2016feedback}, MPID \citep{yang2019bid}; (2) reinforcement learning methods, e.g., USCB \citep{he2021unified}, MAAB \citep{wen2022cooperative}; and (3) generative-model-based methods, e.g., DiffBid \citep{guo2024generative}, GAVE \citep{gao2025generative}.
To cope with the increasingly complex strategies, advertisers begin to use ML models to learn bidding patterns and combine them with human expertise for efficient ad procurement.

\subsection{Bid Optimization by Performance Goal Control}

Following the `Predict, then Optimize' (PO) paradigm \citep{elmachtoub2022smart}, advertisers commonly adopt a two-stage approach for bid optimization. 
The first stage centers on a price-volume model that characterizes how bid price varies with volume (e.g. cost, GMV).
The model learns from historical data with monotonicity constraints as expert priors to improve model reliability. 
The second stage formulates an optimization problem to derive the optimal bidding decision.
PUROS \citep{luo2024puros} models price-volume relationship with monotonic XGBoost model, and computes single-step optimal bids under the `robust satisficing' framework.
GCB-safe \citep{castiglioni2025safe} fits a Gaussian-process price-volume model with smoothness priors, and applies dynamic programming to obtain single-step optimal bids.
Overall, these first-stage models are black-box ML methods learned mostly from historical data, with limited incorporation of expert knowledge. 
Moreover, their second-stage optimization is typically myopic, maximizing single-step reward while ignoring how current bids affect future decisions and rewards, which hinders global revenue maximization.
Under data distribution shifts caused by sales promotions or new product launches, due to the mismatch between online and historical data distributions, price-volume models exhibit degraded performance.
The resulting prediction errors propagate into the second stage, amplifying the unreliability of bidding decisions and potentially causing losses for advertisers.

\subsection{Dual-process Theory in Industrial Applications}

Industrial implementations of dual-process theory adopt two complementary strategies to instantiate System 1 and System 2: System 1 provides robust, low-latency decisions and learned priors to reduce the cost of acquiring complex policies from historical data, while System 2 applies deliberative processing to improve overall performance. 
In autonomous driving, LeapAD \citep{mei2024continuously} pairs an LLM-based System 2 for rule-grounded reflection and memory curation with a lightweight System 1 that retrieves distilled knowledge for rapid control. 
For robotic manipulation, DP-VLA \citep{han2024dual} employs a large System 2 model to infer task-level intent representations that guide a compact System 1 policy for real-time motor execution. 
In dialogue systems, DPDP \citep{he2024planning} dynamically switches between a fast policy language model (System 1) and an MCTS-based planner (System 2) to balance inference efficiency and planning depth.

\begin{figure*}[t]
  \begin{center}
    \includegraphics[width=\textwidth]{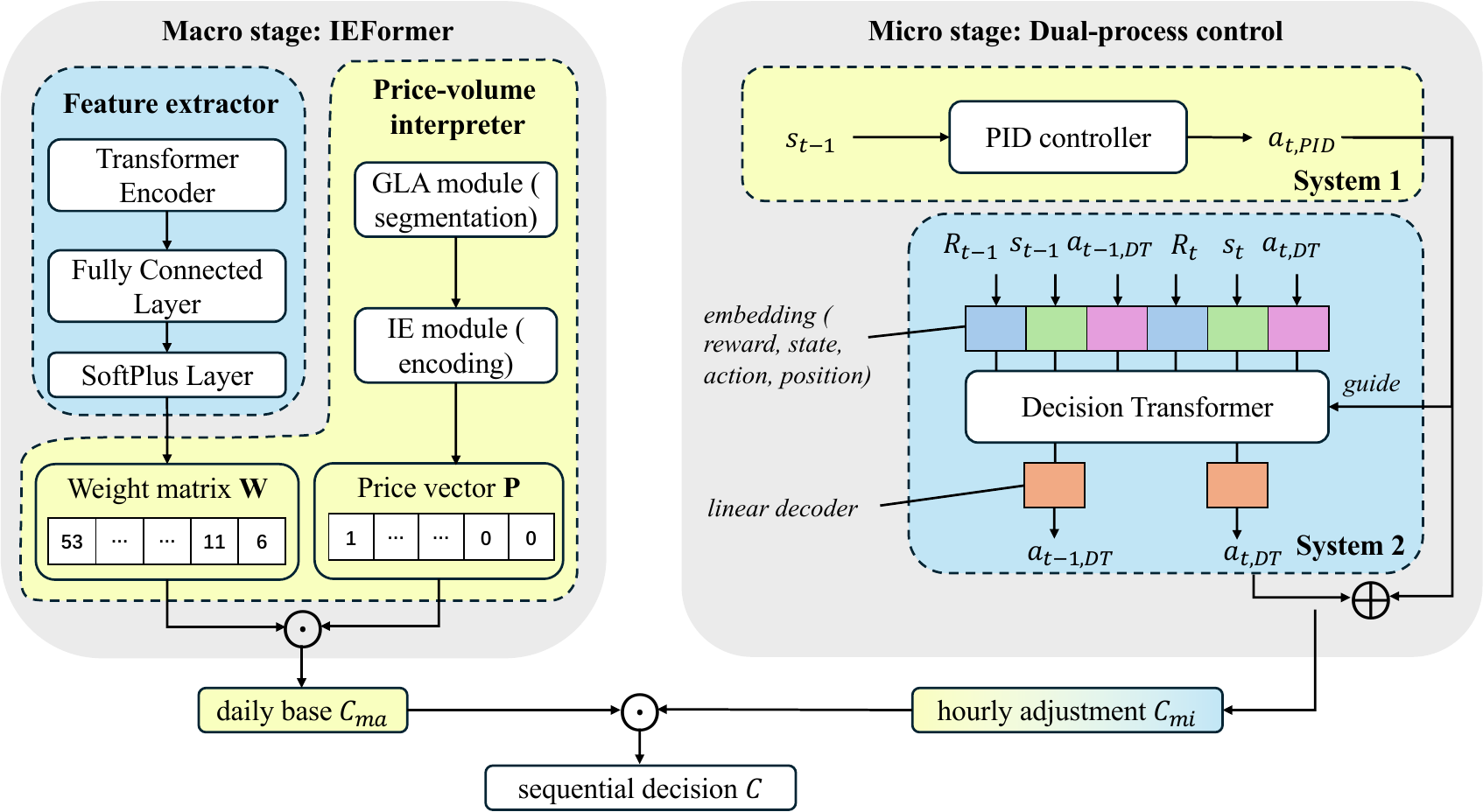}
  \end{center}
  \caption{The architecture of the KBD method. 
  Blue-shaded components denote black-box modules implemented via neural networks, yellow-shaded components indicate rule-based interpretable modules, and mixed-color regions represent their fusion.}
  \label{macro-micro}
\end{figure*}

\section{Problem Formulation}

We consider a sequence of \(M\) ad impressions arriving within a day, indexed by \(i = 1, \ldots, M\). 
Each impression \(i\) is associated with a predicted click-through rate (\(\mathrm{pCTR}_i\)), a predicted conversion rate (\(\mathrm{pCVR}_i\)), a predicted payment amount (\(\mathrm{ppay}_i\)), and a winning price (\(\mathrm{wp}_i\)). 
The advertiser employs the platform-provided tCPA auto-bidding strategy to maximize GMV. 
Under the generalized second-price (GSP) auction mechanism \citep{edelman2007internet}, the strategy submits \(\mathrm{bid}_i\) and wins the impression if \(\mathrm{bid}_i > \mathrm{wp}_i\). 
Let \(x_i\) be the win indicator and \(v_i = \mathrm{pCTR}_i \cdot \text{pCVR}_i \cdot \mathrm{ppay}_i\) the expected conversion value of impression $i$. 
Given budget \(B\) and tCPA target \(C\), the bid optimization problem is formulated as:
\begin{align}
\max_{\{x_i\}} \quad & \sum_{i=1}^{M} x_i \cdot v_i  \\
\text{s.t.} \quad & \sum_{i=1}^{M} x_i \cdot \mathrm{wp}_i \leq B \\
& \frac{\sum_{i=1}^{M} x_i \cdot \mathrm{wp}_i}{\sum_{i=1}^{M} x_i \cdot \text{pCTR}_i \cdot \text{pCVR}_i} \leq C  \\
& x_i \in \{0,1\}, \quad \forall i.
\label{eq:bid_opt_prob}
\end{align}
Following the MPID framework \citep{yang2019bid}, the optimal bid can be expressed as:
\begin{equation}
\mathrm{bid}_i = \frac{1}{p+q} \cdot v_i + \frac{q}{p+q} \cdot \text{pCTR}_i \cdot \text{pCVR}_i \cdot C
\label{eq:optimal_bid}
\end{equation}
where variables \(p, q \ge 0\) correspond to the budget and tCPA constraints respectively, governing spending pace and actual CPA. In practice, \(p\) and \(q\) are tuned by the platform’s auto-bidding system (e.g., GAVE \citep{gao2025generative}), while \(v_i\) and \(\mathrm{ppay}_i\) are estimated via platform ML models.

For conciseness, Eq.~(\ref{eq:optimal_bid}) can be rewritten as:
\begin{equation}
\mathrm{bid}_i = \lambda^0_i + \lambda^1_i \cdot C
\label{eq:optimal_sim_bid}
\end{equation}
where \(\lambda^0_i = \frac{1}{p+q} \cdot v_i\) and \(\lambda^1_i = \frac{q}{p+q}\cdot \text{pCTR}_i \cdot \text{pCVR}_i\) are dynamically set by the platform and unobservable to the advertiser.  
Thus, the advertiser's only control is the tCPA target \(C\), and our objective is to maximize GMV by hourly adjusting \(C\).

\section{Methodology}

As shown in Figure~\ref{macro-micro},  KBD follows the `Predict, then Optimize' framework, and refines the task of tuning parameter C into the following formulation:
\begin{equation}
{C}={C}_{ma} \cdot {C}_{mi}
\label{eq:C}
\end{equation}
where ${C}_{ma}$ is a day-level base tCPA from the macro stage model IEFormer (Isotonic-Embedding based Transformer), and ${C}_{mi}$ is an hour-level adjustment coefficient from the micro stage that combines DT model (System 2) with a PID controller (System 1) to achieve globally optimal bids.

\subsection{Macro stage: IEFormer}

IEFormer learns the tCPA-cost relationship to establish a near-globally optimal tCPA baseline for bid optimization.
Inspired by the Knowledge Integration framework from IML \citep{von2021informed}, we incorporate human expertise into the model at the hypothesis, algorithm and data levels.

\subsubsection{Hypothesis Level}


To reconcile the need for both interpretability (for expert validation and debugging) and expressive power (for capturing complex patterns), we instantiate a hybrid cognitive architecture \citep{mumuni2024improving} that explicitly integrates connectionist learning with symbolic reasoning. Specifically, our architecture comprises two tightly coupled components:

\begin{itemize}
\item A connectionist module: a Transformer encoder \citep{vaswani2017attention} that acts as a feature extractor, distilling historical bidding data into a dense ad embedding. This module learns statistical regularities in an end-to-end manner.
\item A symbolic module: a price-volume interpreter grounded in monotonic piecewise-linear reasoning. It maps cost to tCPA via $ \mathrm{tCPA} = \mathbf{P} \cdot \mathbf{W} $, where $ \mathbf{W} \in \mathbb{R}^N $ represents interpretable marginal tCPA weights over $ N=10 $ cost segments, and $ \mathbf{P} \in [0,1]^N $ is generated by an Isotonic Embedding (IE) module \citep{shen2021framework}.
\end{itemize}

The two modules interact bidirectionally: the connectionist module provides a rich, contextualized representation (the ad embedding) that personalizes the symbolic interpreter’s parameters (\(\textbf{W}\)), while the symbolic structure imposes inductive biases that constrain and regularize the black-box learning process, thereby enabling diagnosis, calibration, and trust.

A key challenge is avoiding skewed interval allocation (e.g., most samples falling into a few segments), which leads to poor estimation in sparse regions.
To address this, we propose an entropy-driven adaptive partitioning strategy: for campaign \( i \) with \( Z_i \) historical samples, let \( z_{ij} \) denote the number assigned to interval \( j \), and define empirical probability \( p_{ij} = z_{ij}/Z_i \). 
We then seek interval boundaries that maximize total information entropy across all campaigns:
\begin{equation}
\max_{S^*} \sum_i \sum_{j=1}^N -p_{ij} \log_N(p_{ij})
\end{equation}
which encourages near-uniform sample distribution and improves generalization in data-sparse regimes. 
This optimization is efficiently solved via the Generalized Lloyd Algorithm (GLA) from data compression literature \citep{chou2002entropy}, repurposed here as a novel mechanism for structuring interpretable cost segments grounded in information-theoretic principles.
Finally, \( \mathbf{W} \) is generated from the ad embedding via a fully connected layer, enabling joint training and campaign-specific personalization.

\subsubsection{Algorithm Level}

In online advertising, domain expertise regarding the tCPA–cost relationship typically asserts two properties: monotonicity, meaning that tCPA is non-decreasing with respect to cost, and smoothness, meaning that the relationship exhibits no abrupt discontinuities.
Through data analysis and discussions with advertising experts, we preserve monotonicity and smoothness, and further introduce a new inductive bias: diminishing marginal returns \citep{wu2022framework}, i.e., the rate of increase in tCPA slows as cost grows.

Monotonicity is implemented by appending a Softplus activation layer after the fully connected layer in the price-volume interpreter, ensuring that \(\mathbf{W}\) is non-negative.  
Smoothness is enforced by adding a smoothness regularizer \(L_\mathrm{{smooth}}\) to the Mean Squared Error (MSE) loss to penalize differences between adjacent elements of \(\mathbf{W}\), thereby discouraging sharp transitions
\begin{equation}
L_{\mathrm{smooth}} = \sum_i \frac{(w_i - w_{i+1})^2}{w_i \cdot w_{i+1}}
\label{eq:smooth_loss}
\end{equation}
To enforce the newly introduced constraint, we introduce a dedicated regularization term \(L_{\mathrm{margin}}\) 
\begin{equation}
L_{\mathrm{margin}} = \sum_i ReLU(\frac{w_{i+1}}{step_{i+1}} - \frac{w_i}{step_i})
\label{eq:margin_loss}
\end{equation}
where \(w_i\) denotes the $i$-th element of the weight vector $\mathbf{W}$ and $step_i$ is the width of the corresponding cost interval. 
This loss penalizes violations of the diminishing marginal returns condition.
We incorporate the marginal utility loss \(L_{\mathrm{margin}}\) into the total objective alongside the prediction loss \(L_{MSE}\) and smoothness regularizer \(L_{\mathrm{smooth}}\)
\begin{equation}
L_{\textrm{IEFormer}} = L_{\textrm{MSE}} + \alpha_{\textrm{smooth}} L_{\textrm{smooth}} + \alpha_{\textrm{margin}} L_{\textrm{margin}}
\label{eq:total_loss}
\end{equation}
where \(\alpha_{\text{smooth}}=0.5, \alpha_{\text{margin}}=0.1\).

\subsubsection{Data Level}

To mitigate data scarcity in online advertising, we transfer bidding knowledge from other auto-bidding strategies. 
Although these strategies optimize for different performance goals (e.g., tROI, tCPC), they all produce bids in the same unit and compete on the same platform.
Leveraging this commonality, we use the platform’s unified eCPM (effective cost-per-mille) formula (Eq.~(\ref{eq:ecpm})) to convert heterogeneous performance goals into equivalent tCPA values \citep{zhang2024optimized}. 
This transformation enables us to enrich the training data for learning the tCPA-cost relationship, effectively augmenting supervision from diverse bidding strategies.
\begin{equation}
\mathbf{tCPA} = \frac{\mathbf{eCPM}}{\text{pCTR} \cdot {\text{pCVR}} \cdot 1000}
= \frac{ppay}{\mathbf{tROI}} 
= \frac{\mathbf{tCPC}}{\mathrm{pCVR}}
\label{eq:ecpm}
\end{equation}
For tROI strategy, we instantiate \(ppay\) in Eq.~(\ref{eq:ecpm}) with the item’s listed price; for tCPC strategy, we instantiate \(\mathrm{pCVR}\) with the campaign’s average conversion rate.  
These augmented samples are used in training with a 0.1 weight to preserve fidelity to native tCPA data.  

\subsection{Micro stage: Dual-process Control with PID \& DT}
At the micro stage, we perform hourly tCPA adjustments to maximize GMV. 
Unlike prior methods that rely on myopic, single-step optimization and ignore inter-temporal dependencies and future ad dynamics, we frame bidding as a sequential decision problem and employ a DT model \citep{chen2021decision} under the offline reinforcement learning paradigm to optimize long-horizon GMV.

\subsubsection{DT for Sequential Bid Optimization}  
We formulate the bid optimization problem as a 24-step Markov decision process (MDP) \(\langle \mathcal{S}, \mathcal{A}, \mathcal{R}, \mathcal{P} \rangle\), where \(s_t \in \mathcal{S}\) denotes the system state at hour \(t\), \(a_t \in \mathcal{A}\) the bidding action, \(r_t\) the immediate reward, and \(\mathcal{P}\) the state transition dynamics. 
The objective is to find an optimal bidding sequence \(a_1, \dots, a_{24}\) that maximizes cumulative return.

In our scenario, state \(s_t\) is a feature vector comprising campaign-level metadata and historical performance statistics (e.g., strategy type, remaining budget);
action \(a_t \in [0.8, 1.2]\) is a multiplicative bid coefficient;
reward \(r_t\) combines normalized cumulative GMV and a spend-utilization penalty:  
  \begin{equation}
  r_t = \frac{\text{GMV}_t}{\text{GMV}^*} + \min\left(0, 1 - \frac{\text{cost}_t}{B}\right)
  \label{eq:step_reward}
  \end{equation}
  where \(\text{GMV}_t\) and \(\text{cost}_t\) are cumulative GMV and spend up to hour \(t\), \(\text{GMV}^*\) is the average GMV over past 7 days.

DT is trained with the following loss \citep{su2024auctionnet}
\begin{equation}
L_{\text{DT}} = L_{\text{MSE}}(a_{\text{t,DT}}, ~ a_{\text{t,gt}})
\label{eq:DT_loss}
\end{equation}  
where \(a_{t,\text{DT}}\) and \(a_{t,\text{gt}}\) denote predicted and ground-truth bid coefficients respectively, and \(a_t\) is the abbreviation of \(a_{t, \text{DT}}\).

\subsubsection{Dual-process Control with PID \& DT}

In real-world deployment, we observed that DT suffers performance degradation during sales promotions or new product launches, where the distributional shift between historical training data and online data renders learned policies ineffective.
Inspired by dual-process theory, we propose to encode expert-derived bidding heuristics into a robust, rule-based controller that acts as System 1 to assist the deliberative DT (System 2) in out-of-distribution scenarios.
Specifically, we integrate a rule-based PID controller (System 1) grounded in expert heuristics that regulates bids based on spending rate deviation
\begin{equation}
a_{\text{t,PID}} = k_p (\text{cost}_{t} - \text{budget}_{t}) + k_i\sum_{hi=1}^{t}(\text{cost}_{hi} - \text{budget}_{hi})
\label{eq:PID}
\end{equation}
with \(k_p, k_i\) weighting current / cumulative cost errors, and \(\text{cost}_t\), \(\text{budget}_t\) the actual / ideal cumulative costs up to hour \(t\).

Drawing on insights from industrial applications of dual-process theory, DT and PID are fused in two dimensions:
(1) during training, we regularize DT toward PID using a Minimum Description Length (MDL) prior  \citep{moskovitz2022unified};
(2) at inference, we dynamically blend decisions based on DT's predictive uncertainty \citep{daw2005uncertainty}.

The MDL framework posits that dual-process architectures emerge from a unified computational objective \citep{grunwald2007minimum}: maximizing reward while minimizing the description length of behavioral policies, i.e., favoring simpler, more compressible strategies.
We thus augment the DT training objective to balance expected return against policy complexity and deviation from the PID baseline, via a regularization term that penalizes excessive divergence from PID
\begin{equation}
L_{\text{DT}} = L_{\text{MSE}}(a_{\text{t,DT}},~a_{t,\text{gt}}) + \beta \cdot L_{\text{MSE}}(a_{t,\text{DT}},~a_{t,\text{PID}})
\label{eq:MDL_loss}
\end{equation}
where $\beta = 0.1$.
At decision time, we employ an uncertainty-weighted fusion mechanism. 
Since the PID controller is deterministic, uncertainty is quantified solely through the DT’s recent prediction error. 
Let \(\mathrm{mape} \ge 0 \) denote the mean absolute percentage error of the DT’s bid predictions over the past three hours, which serves as a proxy for model uncertainty. 
We then compute the final bid coefficient as
\begin{equation}
C_{mi} = \text{max}(1 - \text{mape}, 0) \cdot a_{t,\text{DT}} + \text{min}(\text{mape}, 1) \cdot a_{t,\text{PID}}
\label{eq:fuse}
\end{equation}
This ensures graceful degradation: as DT uncertainty increases, control shifts toward the robust PID policy.

\section{Experiments}
Several experiments are conducted for the following key Research Questions:

\noindent\textbf{RQ1:} Does KBD outperform state-of-the-art methods?

\noindent\textbf{RQ2:} What is the impact of IML on the macro stage of KBD?

\noindent\textbf{RQ3:} Is IEFormer robust to the segment number N?

\noindent\textbf{RQ4:} How does the dual-process control affect the overall performance of KBD?

\begin{table}[H]
\centering
  \caption{Results of click maximization on \textit{iPinYou} dataset.}
  \label{tab1}
  \vskip 0.12in
  \begin{tabular}{p{4.769cm}p{0.8cm}p{1.55cm}}
    \hline
    \multirow{2}{*}{Method}  & \multirow{2}{*}{$R/R^{*}\uparrow$} &constraint satisfaction$\uparrow$ \\ \hline
    ARTEO \citep{korkmaz2022safe}    & 0.601 & 80.65\% \\ 
    BOCO \citep{liang2023online}   & 0.618 & 74.10\% \\ 
    SPB \citep{su2024spending}       & 0.699 & 75.32\% \\
    PUROS \citep{luo2024puros} & 0.723 & 81.73\% \\ 
    GCB-safe \citep{castiglioni2025safe} & 0.606 & 80.77\% \\ 
    \textbf{KBD} w/o DT     & 0.706 & 81.43\% \\ 
    \textbf{KBD} w/o PID    & 0.721 & 80.32\% \\ 
    \hline
    \textbf{KBD}         & \textbf{0.730} & \textbf{82.78\%} \\ \hline
  \end{tabular}
\end{table}

\subsection{Experimental Setup}
\noindent\textbf{Datasets and Evaluation Metrics.}
KBD is evaluated on two datasets: the public \textit{iPinYou} dataset (10-day logs from 9 advertisers, tCPC strategy for click maximization) and a private E-Commerce Advertising dataset (\textit{ECA}) from 200 advertisers (Sep-Dec 2025), optimizing GMV under tCPA / tROI strategies in diverse scenarios including monthly sales promotions and new product launches. 
Following PUROS, we report normalized return $R/R^*$ (actual vs.\ oracle clicks) on \textit{iPinYou}, and model accuracy via wmape (cost-weighted mape) and perf10 (fraction of predictions $\le$ 10\% error) on \textit{ECA}.
Synthetic Control Method \citep{abadie2015comparative} is used in online experiments to evaluate the approach's efficacy.

\textbf{Compared Methods.}
On \textit{iPinYou}\citep{zhang2014real}, we compare KBD against state-of-the-art baselines such as PUROS.
On \textit{ECA}, IEFormer is evaluated against two classes of price–volume models: (1) conventional regressors used in prior work such as GP; and (2) monotonic demand models such as XGBoost.
To avoid harming advertisers' revenue, only methods with PID module are deployed in online tests. 

\subsection{Main Results (RQ1)}
The overall performance of KBD and other baselines  on the \textit{iPinYou} dataset is summarized in Table~\ref{tab1}.
KBD consistently outperforms all prior approaches in both $R/R^{*}$ and constraint satisfaction, providing strong confidence in deploying KBD on real-world ad platforms.
Online tests are conducted on a real-world platform (from which the \textit{ECA} dataset was collected), and results are reported in Table 2.
We observe that IEFormer alone improves cost-exhaust ratio by 8.4\%, while DT contributes an additional 2.3\% improvement.
IEFormer drives the larger gain, underscoring that a well-calibrated price-volume model provides a reliable daily tCPA anchor, which in turn enables more effective hourly bid optimization. 

\begin{table*}[htp]
\centering
  \caption{Online test results on \textit{ECA}: relative lifts against control group; absolute metrics redacted for commercial confidentiality.}
  \label{tab2}
  \vskip 0.12in
  \begin{tabular}{cccccccc}
  \hline
    \multirow{2}{*}{Experimental Period} & \multirow{2}{*}{Group} & \multicolumn{3}{c}{Group Detail}  & \multirow{2}{*}{cost-exhaust ratio} & \multirow{2}{*}{GMV} & \multirow{2}{*}{\makecell{Campaign\\Duration}} \\
    & & use PID & use IEFormer & use DT & & & \\
    \hline
    2025/09/28-2025/10/18 &\makecell{Control\\Treatment} & \makecell{\checkmark \\ \checkmark}  & \makecell{~ \\ \checkmark} & \makecell{~ \\ ~} & \makecell{/\\+~8.44\%} & \makecell{/\\+~6.14\%} &\makecell{/\\+~0.58\%} \\ \hline
    2025/11/20-2025/12/10 &\makecell{Control\\Treatment} & \makecell{\checkmark \\ \checkmark}  & \makecell{~ \\ \checkmark} & \makecell{~ \\ \checkmark}  & \makecell{/\\+14.55\%} & \makecell{/\\+13.01\%} & \makecell{/\\+1.36\%} \\ \hline
    2025/12/11-2025/12/31 &\makecell{Control\\Treatment} & \makecell{\checkmark \\ \checkmark}  & \makecell{\checkmark \\ $\checkmark$} & \makecell{~ \\ \checkmark}  & \makecell{/\\+2.27\%} & \makecell{/\\+2.66\%} & \makecell{/\\+0.20\%} \\ \hline
  \end{tabular}
\end{table*}

\begin{figure}[!htp]
  \begin{center}
    \includegraphics[width=\columnwidth]{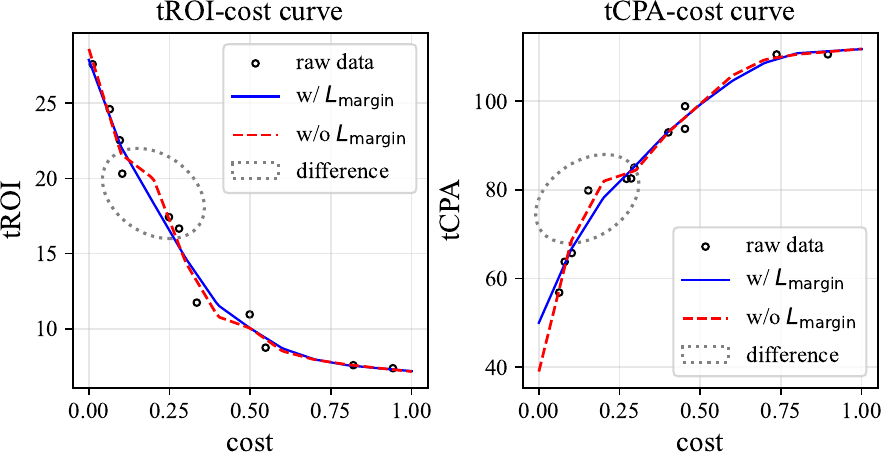}
  \end{center}
  \caption{Case studies on the influence of $L_{\mathrm{margin}}$. 
  As highlighted by the ellipsoidal regions, the price-volume curve is more prone to overfitting the noisy data without $L_{\mathrm{margin}}$.}
  \label{curve-figure}
\end{figure}

\subsection{Further Analysis}

\subsubsection{Ablation Study of IML (RQ2)}
Ablation studies on the \textit{ECA} dataset are conducted to assess the contribution of knowledge at each level of IML framework. 
Specifically, we consider four variants:
(1) \textit{w/o IE}: replaces IE module with end-to-end prediction;
(2) \textit{w/o GLA}: replaces GLA module with fixed segmentation;
(3) \textit{w/o $L_{\textrm{margin}}$ }: disables the $L_{\textrm{margin}}$ loss;
(4) \textit{w/o tROI data}: removes training samples from tROI strategies.

As shown in Table~\ref{tab3}, all ablated variants suffer immediate performance degradation, confirming that knowledge integration across all levels is essential. 
Notably, removing IE module leads to the largest drop, highlighting the critical role of the price-volume interpreter in the hybrid cognitive architecture.

\begin{table}[H]
\centering
  \caption{Ablation Study of IML on \textit{ECA} dataset.}
  \label{tab3}
  \vskip 0.12in
  \begin{tabular}{lcccc}
    \hline
    Method  & wmape $\downarrow$ & mape $\downarrow$ & perf10 $\uparrow$ \\ \hline
    w/o IE    & 0.2588 & 0.2348 &  0.2738 \\ 
    w/o GLA    & 0.2412 & 0.2270 & 0.2793 \\
    w/o $L_{\mathrm{margin}}$ & 0.2386 & 0.2268 & 0.2921 \\ 
    w/o tROI data & 0.2226 & 0.2600 & 0.2976 \\
    \hline
    \textbf{KBD}    & \textbf{0.2187} & \textbf{0.2151} & \textbf{0.3214} \\ 
    \hline
  \end{tabular}
\end{table}

Moreover, ablating $L_{\mathrm{margin}}$ also harms performance.
To illustrate its influence, we conduct a case study using the \textbf{W} matrix 
from the price-volume interpreter to visualize the learned price-volume curves. 
As Figure~\ref{curve-figure} shows, $L_{\mathrm{margin}}$ acts as an effective regularizer. 
It prevents overfitting to noisy data and guides the model to produce monotonic, smooth curves that align with expert expectations of diminishing returns.

Table~\ref{tab4} shows experiments that benchmark mainstream price-volume models on the \textit{ECA} dataset. 
Black-box approaches such as IGCM \citep{tang2025learning}, despite enforcing monotonicity, perform poorly on \textit{ECA}. 
This underscores the necessity of using IE module as an explicit, interpretable price-volume interpreter when applying deep learning in such dynamic environments. 
Additionally, although DIPN \citep{shen2021framework} also employs IE module, it still underperforms IEFormer markedly, indicating that the powerful feature extractor is equally crucial in our hybrid cognitive architecture. 

\begin{table}[H]
\centering
  \caption{Price-volume model performance on \textit{ECA} dataset.}
  \label{tab4}
  \vskip 0.12in
  \begin{tabular}{lccc}
    \hline
    Method  & wmape $\downarrow$ & mape $\downarrow$ \\ 
    \hline
    IGCM \citep{tang2025learning} & 0.9979 & 0.9949 \\
    GP \citep{schulz2018tutorial}    & 0.6414 & 1.8021  \\ 
    CMNN \citep{li2024enhancing}    & 0.6309  & 2.0599 \\
    DIPN \citep{shen2021framework} & 0.6308 & 1.7647 \\
    SMNN \citep{kim2024scalable} & 0.6179 & 2.2503 \\
    SMM \citep{igel2023smooth} & 0.6023 & 1.6785 \\
    XGBoost \citep{chen2016xgboost}    & 0.4967 & 1.1872 \\ 
    \hline
    \textbf{IEFormer} & \textbf{0.2187} & \textbf{0.2151} \\
    \hline
  \end{tabular}
\end{table}

\subsubsection{Robustness of IEFormer to Number N (RQ3)}
\begin{figure}[H]
  \begin{center}
    \includegraphics[width=\columnwidth]{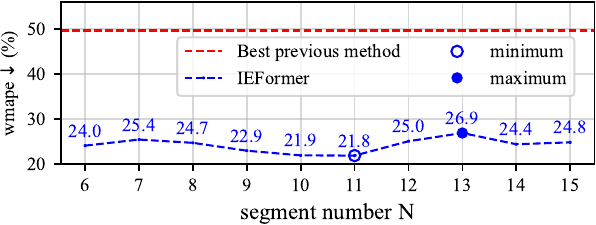}
  \end{center}
  \caption{Robustness of IEFormer to segment number N.}
  \label{wmape-curve}
\end{figure}

To evaluate the robustness of IEFormer to the number of segments N, we vary N from 6 to 15 and measure its impact on model performance. 
As shown in Figure~\ref{wmape-curve}, although wmape fluctuates between 21\% and 27\%, IEFormer consistently outperforms all baseline price-volume models in Table~\ref{tab4}, demonstrating strong robustness of the IE module to the choice of 
N.

\subsubsection{Ablation Study of Dual-process Control (RQ4)}
On \textit{iPinYou} dataset, we compare the standalone DT, PID controller, and their integration via dual-process control.
As shown in Table~\ref{tab1}, DT improves \(R/R^*\) by optimizing sequential decisions globally at the cost of reduced constraint satisfaction, since its System 2–like reasoning occasionally produces aggressive bids that violate advertiser constraints. 
In contrast, the PID controller, a System 1 rule-based policy, ensures reliable constraint adherence across all cases but sacrifices potential clicks due to its myopic, stepwise optimality.  

The dual-process control framework reconciles these trade-offs and improves both metrics by roughly 1\%: 
by incorporating PID’s decisions as a soft constraint during DT training, DT model learns to emulate PID’s conservative behavior, reducing constraint violations. 
Simultaneously, DT calibrates PID’s bids to maximize long-term reward, yielding higher \(R/R^*\) than either component alone. 

We observe consistent gains in online experiments on the \textit{ECA} dataset (Table~\ref{tab2}). 
The dual-process control framework significantly improves cost-exhaust ratio, GMV and campaign duration over the PID baseline, demonstrating its effectiveness in real-world scenarios involving distribution shifts. 
The framework generalizes across multiple auto-bidding strategies, confirming that integrating fast, robust heuristics with slow, deliberative planning yields more adaptive and reliable decision-making in dynamic advertising environments.

\section{Conclusion}  
We propose KBD, a two-stage bidding framework that integrates expert knowledge into the macro-stage price–volume model via IML and optimizes long-horizon returns in the micro stage using DT. 
Drawing on dual-process theory, we fuse DT (System 2) with a PID controller (System 1) to balance deliberative planning and reactive control.
Experiments on \textit{iPinYou} and a real-world platform show that KBD outperforms state-of-the-art methods, with gains attributed to richer knowledge infusion and dual-process robustness under real-world distribution shifts.
Future work will leverage large language models to build bidding agents endowed with human-like cognitive mechanisms, capable of optimal decisions and generating causal, goal-directed natural-language explanations to render its reasoning transparent and interpretable.

\printbibliography

@report{emarketer2025programmaticH1,
  author      = {Evelyn Mitchell-Wolf},
  title       = {Programmatic Advertising Forecast and Ad Tech Trends H1 2025},
  institution = {EMARKETER Inc.},
  date        = {2025-02-12},
  type        = {Report}
}

@article{liang2023online,
  title={Online ad procurement in non-stationary autobidding worlds},
  author={Liang, Jason Cheuk Nam and Lu, Haihao and Zhou, Baoyu},
  journal={Advances in Neural Information Processing Systems},
  volume={36},
  pages={42552--42575},
  year={2023}
}

@inproceedings{luo2024puros,
  title={PUROS: A Cpx-Ievered Framework for Ad Procurement in Autobidding Worlds},
  author={Luo, Huixiang and Gao, Longyu and Huang, Pingchun and Li, Tianning},
  booktitle={2024 IEEE 13th Data Driven Control and Learning Systems Conference (DDCLS)},
  pages={1399--1406},
  year={2024},
  organization={IEEE}
}

@inproceedings{castiglioni2025safe,
  title={Safe Online Bid Optimization with Return on Investment and Budget Constraints},
  author={Castiglioni, Matteo and Nuara, Alessandro and Romano, Giulia and Spadaro, Giorgio and Trov{\`o}, Francesco and Gatti, Nicola},
  booktitle={Proceedings of the 31st ACM SIGKDD Conference on Knowledge Discovery and Data Mining V. 1},
  pages={73--81},
  year={2025}
}

@article{von2021informed,
  title={Informed machine learning--a taxonomy and survey of integrating prior knowledge into learning systems},
  author={Von Rueden, Laura and Mayer, Sebastian and Beckh, Katharina and Georgiev, Bogdan and Giesselbach, Sven and Heese, Raoul and Kirsch, Birgit and Pfrommer, Julius and Pick, Annika and Ramamurthy, Rajkumar and others},
  journal={IEEE Transactions on Knowledge and Data Engineering},
  volume={35},
  number={1},
  pages={614--633},
  year={2021},
  publisher={IEEE}
}

@article{mumuni2024improving,
  title={Improving deep learning with prior knowledge and cognitive models: A survey on enhancing explainability, adversarial robustness and zero-shot learning},
  author={Mumuni, Fuseini and Mumuni, Alhassan},
  journal={Cognitive Systems Research},
  volume={84},
  pages={101188},
  year={2024},
  publisher={Elsevier}
}

@article{chen2021decision,
  title={Decision transformer: Reinforcement learning via sequence modeling},
  author={Chen, Lili and Lu, Kevin and Rajeswaran, Aravind and Lee, Kimin and Grover, Aditya and Laskin, Misha and Abbeel, Pieter and Srinivas, Aravind and Mordatch, Igor},
  journal={Advances in neural information processing systems},
  volume={34},
  pages={15084--15097},
  year={2021}
}

@article{moskovitz2022unified,
  title={A unified theory of dual-process control},
  author={Moskovitz, Ted and Miller, Kevin and Sahani, Maneesh and Botvinick, Matthew M},
  journal={arXiv preprint arXiv:2211.07036},
  year={2022}
}

@article{bennett1993development,
  title={Development of the PID controller},
  author={Bennett, Stuart},
  journal={IEEE Control Systems Magazine},
  volume={13},
  number={6},
  pages={58--62},
  year={1993},
  publisher={IEEE}
}

@inproceedings{yang2019bid,
  title={Bid optimization by multivariable control in display advertising},
  author={Yang, Xun and Li, Yasong and Wang, Hao and Wu, Di and Tan, Qing and Xu, Jian and Gai, Kun},
  booktitle={Proceedings of the 25th ACM SIGKDD international conference on knowledge discovery \& data mining},
  pages={1966--1974},
  year={2019}
}

@article{weinan2016feedback,
  title={Feedback control of real-time display advertising},
  author={Weinan, W and Rong, Y and Wang, J and Zhu, T and Wang, X},
  journal={Proceedings of the Web Search and Data Mining (WSDM)},
  pages={407--416},
  year={2016}
}

@inproceedings{he2021unified,
  title={A unified solution to constrained bidding in online display advertising},
  author={He, Yue and Chen, Xiujun and Wu, Di and Pan, Junwei and Tan, Qing and Yu, Chuan and Xu, Jian and Zhu, Xiaoqiang},
  booktitle={Proceedings of the 27th ACM SIGKDD Conference on Knowledge Discovery \& Data Mining},
  pages={2993--3001},
  year={2021}
}

@inproceedings{wen2022cooperative,
  title={A cooperative-competitive multi-agent framework for auto-bidding in online advertising},
  author={Wen, Chao and Xu, Miao and Zhang, Zhilin and Zheng, Zhenzhe and Wang, Yuhui and Liu, Xiangyu and Rong, Yu and Xie, Dong and Tan, Xiaoyang and Yu, Chuan and others},
  booktitle={Proceedings of the Fifteenth ACM International Conference on Web Search and Data Mining},
  pages={1129--1139},
  year={2022}
}

@inproceedings{guo2024generative,
  title={Generative auto-bidding via conditional diffusion modeling},
  author={Guo, Jiayan and Huo, Yusen and Zhang, Zhilin and Wang, Tianyu and Yu, Chuan and Xu, Jian and Zheng, Bo and Zhang, Yan},
  booktitle={Proceedings of the 30th ACM SIGKDD Conference on Knowledge Discovery and Data Mining},
  pages={5038--5049},
  year={2024}
}

@inproceedings{gao2025generative,
  title={Generative auto-bidding with value-guided explorations},
  author={Gao, Jingtong and Li, Yewen and Mao, Shuai and Jiang, Peng and Jiang, Nan and Wang, Yejing and Cai, Qingpeng and Pan, Fei and Jiang, Peng and Gai, Kun and others},
  booktitle={Proceedings of the 48th International ACM SIGIR Conference on Research and Development in Information Retrieval},
  pages={244--254},
  year={2025}
}

@article{mei2024continuously,
  title={Continuously learning, adapting, and improving: A dual-process approach to autonomous driving},
  author={Mei, Jianbiao and Ma, Yukai and Yang, Xuemeng and Wen, Licheng and Cai, Xinyu and Li, Xin and Fu, Daocheng and Zhang, Bo and Cai, Pinlong and Dou, Min and others},
  journal={arXiv preprint arXiv:2405.15324},
  year={2024}
}

@article{han2024dual,
  title={A dual process vla: Efficient robotic manipulation leveraging vlm},
  author={Han, ByungOk and Kim, Jaehong and Jang, Jinhyeok},
  journal={arXiv preprint arXiv:2410.15549},
  year={2024}
}

@article{he2024planning,
  title={Planning like human: A dual-process framework for dialogue planning},
  author={He, Tao and Liao, Lizi and Cao, Yixin and Liu, Yuanxing and Liu, Ming and Chen, Zerui and Qin, Bing},
  journal={arXiv preprint arXiv:2406.05374},
  year={2024}
}

@article{elmachtoub2022smart,
  title={Smart “predict, then optimize”},
  author={Elmachtoub, Adam N and Grigas, Paul},
  journal={Management Science},
  volume={68},
  number={1},
  pages={9--26},
  year={2022},
  publisher={INFORMS}
}

@article{daw2005uncertainty,
  title={Uncertainty-based competition between prefrontal and dorsolateral striatal systems for behavioral control},
  author={Daw, Nathaniel D and Niv, Yael and Dayan, Peter},
  journal={Nature neuroscience},
  volume={8},
  number={12},
  pages={1704--1711},
  year={2005},
  publisher={Nature Publishing Group US New York}
}

@article{vaswani2017attention,
  title={Attention is all you need},
  author={Vaswani, Ashish and Shazeer, Noam and Parmar, Niki and Uszkoreit, Jakob and Jones, Llion and Gomez, Aidan N and Kaiser, {\L}ukasz and Polosukhin, Illia},
  journal={Advances in neural information processing systems},
  volume={30},
  year={2017}
}

@article{shen2021framework,
  title={A framework for massive scale personalized promotion},
  author={Shen, Yitao and Wang, Yue and Lu, Xingyu and Qi, Feng and Yan, Jia and Mu, Yixiang and Yang, Yao and Peng, YiFan and Gu, Jinjie},
  journal={arXiv preprint arXiv:2108.12100},
  year={2021}
}

@article{chou2002entropy,
  title={Entropy-constrained vector quantization},
  author={Chou, Philip A and Lookabaugh, Tom and Gray, Robert M},
  journal={IEEE Transactions on acoustics, speech, and signal processing},
  volume={37},
  number={1},
  pages={31--42},
  year={2002},
  publisher={IEEE}
}

@inproceedings{wu2022framework,
  title={A framework for multi-stage bonus allocation in meal delivery platform},
  author={Wu, Zhuolin and Wang, Li and Huang, Fangsheng and Zhou, Linjun and Song, Yu and Ye, Chengpeng and Nie, Pengyu and Ren, Hao and Hao, Jinghua and He, Renqing and others},
  booktitle={Proceedings of the 28th ACM SIGKDD Conference on Knowledge Discovery and Data Mining},
  pages={4195--4203},
  year={2022}
}

@inproceedings{zhang2024optimized,
  title={Optimized Cost Per Click in Online Advertising: A Theoretical Analysis},
  author={Zhang, Kaichen and Yuan, Zixuan and Xiong, Hui},
  booktitle={Proceedings of the 30th ACM SIGKDD Conference on Knowledge Discovery and Data Mining},
  pages={4232--4243},
  year={2024}
}

@article{su2024auctionnet,
  title={Auctionnet: A novel benchmark for decision-making in large-scale games},
  author={Su, Kefan and Huo, Yusen and Zhang, Zhilin and Dou, Shuai and Yu, Chuan and Xu, Jian and Lu, Zongqing and Zheng, Bo},
  journal={Advances in Neural Information Processing Systems},
  volume={37},
  pages={94428--94452},
  year={2024}
}

@book{grunwald2007minimum,
  title={The minimum description length principle},
  author={Gr{\"u}nwald, Peter D},
  year={2007},
  publisher={MIT press}
}

@article{abadie2015comparative,
  title={Comparative politics and the synthetic control method},
  author={Abadie, Alberto and Diamond, Alexis and Hainmueller, Jens},
  journal={American Journal of Political Science},
  volume={59},
  number={2},
  pages={495--510},
  year={2015},
  publisher={Wiley Online Library}
}

@article{li2024enhancing,
  title={Enhancing Monotonic Modeling with Spatio-Temporal Adaptive Awareness in Diverse Marketing},
  author={Li, Bin and Pei, Jiayan and Xiao, Feiyang and Zhao, Yifan and Zhang, Zhixing and Liu, Diwei and He, HengXu and Jia, Jia},
  journal={arXiv preprint arXiv:2406.14132},
  year={2024}
}

@inproceedings{su2024spending,
  title={Spending programmed bidding: Privacy-friendly bid optimization with roi constraint in online advertising},
  author={Su, Yumin and Xiang, Min and Chen, Yifei and Li, Yanbiao and Qin, Tian and Zhang, Hongyi and Li, Yasong and Liu, Xiaobing},
  booktitle={Proceedings of the 30th ACM SIGKDD Conference on Knowledge Discovery and Data Mining},
  pages={5731--5740},
  year={2024}
}

@article{korkmaz2022safe,
  title={Safe and adaptive decision-making for optimization of safety-critical systems: The ARTEO algorithm},
  author={Korkmaz, Buse Sibel and Zag{\'o}rowska, Marta and Mercang{\"o}z, Mehmet},
  journal={arXiv preprint arXiv:2211.05495},
  year={2022}
}

@article{schulz2018tutorial,
  title={A tutorial on Gaussian process regression: Modelling, exploring, and exploiting functions},
  author={Schulz, Eric and Speekenbrink, Maarten and Krause, Andreas},
  journal={Journal of mathematical psychology},
  volume={85},
  pages={1--16},
  year={2018},
  publisher={Elsevier}
}

@article{chen2016xgboost,
  title={XGBoost: A Scalable Tree Boosting System},
  author={Chen, Tianqi},
  journal={Cornell University},
  year={2016}
}

@article{igel2023smooth,
  title={Smooth min-max monotonic networks},
  author={Igel, Christian},
  journal={arXiv preprint arXiv:2306.01147},
  year={2023}
}

@inproceedings{kim2024scalable,
  title={Scalable monotonic neural networks},
  author={Kim, Hyunho and Lee, Jong-Seok},
  booktitle={The Twelfth International Conference on Learning Representations},
  year={2024}
}

@article{tang2025learning,
  title={Learning Monotonic Probabilities with a Generative Cost Model},
  author={Tang, Yongxiang and Cheng, Yanhua and Liu, Xiaocheng and Jiao, Chenchen and Zeng, Yanxiang and Luo, Ning and Yuan, Pengjia and Liu, Xialong and Jiang, Peng},
  journal={arXiv preprint arXiv:2506.03542},
  year={2025}
}

@article{edelman2007internet,
  title={Internet advertising and the generalized second-price auction: Selling billions of dollars worth of keywords},
  author={Edelman, Benjamin and Ostrovsky, Michael and Schwarz, Michael},
  journal={American economic review},
  volume={97},
  number={1},
  pages={242--259},
  year={2007},
  publisher={American Economic Association}
}

@article{zhang2014real,
  title={Real-time bidding benchmarking with ipinyou dataset},
  author={Zhang, Weinan and Yuan, Shuai and Wang, Jun and Shen, Xuehua},
  journal={arXiv preprint arXiv:1407.7073},
  year={2014}
}

\end{document}